\documentclass[runningheads]{llncs}
\usepackage[T1]{fontenc}
\usepackage{graphicx}
\usepackage{color}

\usepackage{multirow}%
\usepackage{amsmath,amsfonts}
\usepackage{algorithmic}
\usepackage{algorithm}
\usepackage{array}
\usepackage[caption=false,font=normalsize,labelfont=sf,textfont=sf]{subfig}
\usepackage{textcomp}
\usepackage{stfloats}
\usepackage{url}
\usepackage{verbatim}
\usepackage{cite}
\usepackage{hyperref}

\newcommand{\x}{\mathbf{x}}
\newcommand{\z}{\mathbf{z}}

\newcommand{\zero}{\mathbf{0}}
\newcommand{\Id}{\mathbf{I}}
\newcommand{\R}{\mathbb{R}}
\newcommand{\Gauss}[2]{\mathcal{N}\left(#1,#2\right)}

\newcommand{\K}[1]{K\left(#1\right)}

\newcommand{\lt}{\ensuremath <}
\newcommand{\gt}{\ensuremath >}

\allowdisplaybreaks
\begin{document}
\title{Matching Aggregate Posteriors in the Variational Autoencoder}
\titlerunning{Matching Aggregate Posteriors in the Variational Autoencoder}

\author{Surojit Saha\inst{1}\orcidID{0000-0003-0198-0189} \and
Sarang Joshi\inst{1} \and
Ross Whitaker\inst{1}}

\authorrunning{S. Saha et al.}

\institute{Scientific Computing and Imaging Institute, The University of Utah, USA.\\
\email{surojit.saha@utah.edu; sjoshi@sci.utah.edu; whitaker@cs.utah.edu}}

\maketitle              
\begin{abstract}
The variational autoencoder (VAE) \cite{kingma2014auto} is a well-studied, deep, latent-variable model (DLVM) that optimizes the variational lower bound of the log marginal data likelihood. However, the VAE's known failure to match the aggregate posterior often results unacceptable latent distribution, e.g. with \emph{pockets, holes, or clusters}, that fail to adequately resemble the prior. The training of the VAE under different scenarios can also result in \emph{posterior collapse}, which is associated with a loss of information in the latent space. This paper addresses these shortcomings in VAEs by reformulating the objective function to match the \emph{aggregate/marginal} posterior distribution to the prior. We use \emph{kernel density estimate} (KDE) to model the aggregate posterior. We propose an automated method to estimate the kernel and account for the associated kernel bias in our estimation, which enables the use of KDE in high-dimensional latent spaces. The proposed method is named the \emph{aggregate variational autoencoder} (AVAE) and is built on the theoretical framework of the VAE. 
Empirical evaluation of the proposed method on multiple benchmark datasets demonstrates the advantages of the AVAE relative to state-of-the-art (SOTA) DLVM methods. Here is the link to the code: \url{https://github.com/Surojit-Utah/AVAE}.

\keywords{Variational Autoencoders \and Aggregate Posterior Matching \and Non-parametric Density Estimation.}
\end{abstract}
\section{Introduction}
The development of DLVMs is an important topic of research that is widely used for generative modeling and representation learning. The VAE  \cite{rezende2014stochastic,kingma2014auto}, a DLVM, learns a joint distribution distribution, $p_{\theta}(\x{}, \z{})$, that captures the relationship between a set of hidden variables, $\z{}$, and the observed variables, $\x{}$. VAEs model the data distribution, $p_{\theta}(\x{}) = \int p_{\theta}(\x{}, \z{}) d\z{} = \int p_{\theta}(\x{} \mid \z{}) p(\z{}) d\z{}$, by optimizing the parameters, $\theta$, typically a deep neural network known as the \emph{generative model/decoder}. The VAE approximates the true posterior by a \emph{surrogate} distribution, $q_{\phi}(\z{} \mid \x{})$, that informs the objective function to use a latent subspace that is likely to maximize $p_{\theta}(\x{} \mid \z{})$. The parameters ($\phi$) of the surrogate posterior is another deep neural network known as the \emph{inference model/encoder}. The encoder ($\phi$) and decoder ($\theta$) parameters of the VAE are jointly optimized to maximize the evidence lower bound (ELBO).

Despite strong theoretical foundations, the VAE fails in matching the \emph{aggregate} posterior $q_{\phi}(\z{}) = \int q_{\phi}(\z{} \mid \x{}) p(\x{}) d\x{}$ to the prior $p(\z{}) = \Gauss{\zero}{\Id}$. The mismatch between distributions results in \emph{clusters} or \emph{holes} in the latent space, indicating regions strongly supported under the prior may have low density under aggregate posterior \cite{ELBO_surgery_2016,rosca2018distribution} (and vice versa). The presence of \emph{holes} increases the mismatch between the learned, $p_{\theta}(\x{})$, and real data distribution, $p(\x{})$, leading to the generation of low-quality samples. To alleviate this issue, methods in \cite{Flex_VAE_Prior_2018,resample_prior_VAE} use a flexible prior, and the authors of \cite{dai2019_2sVAE} match the prior in two stages. An additional regularization loss is added to the ELBO to match aggregate distributions that help in learning meaningful representations \cite{Info_VAE_AAAI_2019} and improved disentanglement of latent factors \cite{Factor_VAE_ICML_2018}. The estimation of $q_{\phi}(\z{})$ is challenging, thus the methods in \cite{Info_VAE_AAAI_2019,Factor_VAE_ICML_2018} uses an adversarial classifier or a kernel method, e.g., MMD \cite{MMD_2012}, to match the aggregate posterior to the prior, similar to the Wasserstein autoencoders \cite{tolstikhin2017wasserstein}.

The posterior distribution of the VAE might collapse to the prior for \emph{a subset} or \emph{all} of the latent dimensions during the training of VAEs. Under such scenarios, the representations produced by the encoder on the collapsed latent dimensions lack information that the decoder can use to reconstruct the input faithfully. This phenomenon is known as the \emph{posterior collapse} or the \emph{KL vanishing effect} \cite{Cycling_KL_annealing_VAE_NAACL_2019,posterior_collapse_delta_VAE_ICLR_2019,posterior_collapse_delta_VAE_ICLR_2019}. We expect to encounter such degenerate solutions more often with the $\beta$-VAE \cite{beta_VAE_ICLR_2017} that advocates the use of higher $\beta$ values for the improved disentanglement of the latent factors. The analysis in \cite{ELBO_surgery_2016} explains the minimization of the mutual information $I(\x{};\z{})$ between the latent ($\z{}$) and observed variables ($\x{}$) for higher $\beta$ values. Several methods have been proposed to circumvent this issue, such as the KL annealing strategy \cite{KL_annealing_VAE_ICML_2016,Cycling_KL_annealing_VAE_NAACL_2019}, explicit inhibition of the distribution matching \cite{posterior_collapse_delta_VAE_ICLR_2019}, use of complex priors \cite{Flex_VAE_Prior_2018,resample_prior_VAE}, and special training policy \cite{Train_VAE_Lag_ICLR_2019}.

In this work, we address the limitations of the VAE by matching the aggregate posterior to the prior in the ELBO framework derived from first principles. We use KDE in the latent space to model the aggregate posterior, $q_{\phi}(\z{})$. The use of KDE in the AVAE helps in a better estimate of differences between distributions relative to the adversarial training and kernel-based method used in \cite{Info_VAE_AAAI_2019,Factor_VAE_ICML_2018,makhzani2016adversarial,tolstikhin2017wasserstein}. In addition to improvement in the quality of the generated samples, matching the aggregate posterior to a prior finds potential application in the meaningful interpretation of the latent generative factors \cite{Factor_VAE_ICML_2018, TC_VAE_Neurips_2019}, outlier detection \cite{gens_saha_2022}, and data completion \cite{Inpaint_VAE_ICLR_2022, Inpaint_VAE_CVPR_2021}. Unlike other variants of the VAE that strive to match marginal posterior to the prior \cite{Factor_VAE_ICML_2018,TC_VAE_Neurips_2019,Info_VAE_AAAI_2019}, the proposed method does not require additional regularization terms or hyperparameters to the objective function. Moreover, we propose a heuristic that automatically adjusts the $\beta$ value during the training instead of empirically estimating the $\beta$ for a dataset using cross-validation. The potential benefits of using KDEs for matching distributions have been thoroughly studied in \cite{gens_saha_2022}. Though KDEs are used in \cite{gens_saha_2022} for matching the aggregate posterior distribution to the prior, the objective function is not derived in the general framework of DLVMs, and it is not well suited to high-dimensional latent spaces, e.g., $\geq 50$, which restricts its application for modeling complex datasets, such as the CIFAR10 \cite{CIFAR10}. We correct the bias in KDE bandwidth estimation that qualifies the AVAE to use KDE in high-dimensional latent spaces (dimensions > 100).

The main contributions of this work are summarized as follows:
\begin{itemize}
  \item Matching the aggregate posterior distribution to the prior in the VAE objective function using KDE without any modification of the ELBO.
  \item An automated method for estimating KDE bandwidth that allows using KDEs in high-dimensional latent spaces (dimensions $\gt 100$).   
  \item Evaluations showing that the AVAE addresses the shortcomings in the formulation (\emph{pockets/clusters}) and training (\emph{posterior collapse}) of the VAE.
  \item The regularization scalar $\beta$ is updated during training using the proposed heuristic. Thus, the AVAE is free from tuning the hyperparameter, $\beta$.
  \item Empirical evaluation of the proposed method using different efficacy measures on multiple benchmark datasets, producing results that compare favorably with state-of-the-art, likelihood-based, generative models.
\end{itemize}

\section{Related Work}\label{sec:related_work}
Several extensions to the formulation of the VAE address known limitations, such as alleviating posterior collapse \cite{posterior_collapse_VQ_VAE_2017_NIPS, posterior_collapse_delta_VAE_ICLR_2019}, better matching of marginal posteriors \cite{Factor_VAE_ICML_2018, TC_VAE_Neurips_2019}, and reducing over-regularization \cite{Flex_VAE_Prior_2018, resample_prior_VAE}. Methods matching marginal posteriors are relevant to our work. These methods introduced an additional regularization term to the objective function \cite{Factor_VAE_ICML_2018, TC_VAE_Neurips_2019} (along with a hyperparameter) to encourage statistical independence of latent factors. An interesting analysis of the VAE objective is done in \cite{RAE} (RAE), which suggests that an autoencoder with regularized decoder parameters is as good as the VAE.

The generative adversarial network (GAN) is another popular generative model that implicitly matches distributions using a discriminator \cite{goodfellow2014generative, dcgan_2016}. GANs produce novel, realistic examples, such as images with sharp and distinct features, which are difficult for even humans to identify as generated images \cite{Style_GAN_2021}. Nevertheless, GANs do not produce a reliable matching form data samples into the latent space \cite{donahue2017adversarial}, and there are significant challenges in optimizing the objective function of a GAN \cite{Conv_GANs_2017_NIPS, Conv_GANs_2018_ICML, Conv_GANs_2018_Oxford}. GANs are very particular about the architecture of the discriminator, training strategy, and the associated hyperparameters \cite{salimans2016improved, GANS_regularized_2017}. The adversarial autoencoder (AAE) \cite{makhzani2016adversarial} is a likelihood-based generative model that implicitly matches the aggregate posterior in the latent space of an autoencoder to a prior with the help of a discriminator.

WAEs \cite{tolstikhin2017wasserstein} is another likelihood-based generative model that explicitly matches the aggregate posterior to a prior in the latent space (unlike VAEs). In the WAE, the Wasserstein distance between the data and generated distribution is minimized by factoring the latent variable $\z{}$ in its formulation. The regularization term in WAEs is computed using two different strategies. In one approach, a discriminator is used in the latent space, as in AAEs, and is known as the WAE-GAN. In the other approach, the maximum mean discrepancy (MMD) \cite{MMD_2012} is used to compute the divergence between distributions in the latent space, known as the WAE-MMD. 

\section{Method}\label{sec:method_headings}
\subsection{Background}\label{sec:method_background}
The goal of a DLVM is to learn the joint distribution of the latent variables, $\z{}$, and the observed variables, $\x{}$, such that the resulting (generative) distribution closely approximates the true but unknown data distribution, $p_{data}(\x{})$.
In the DLVM, $p_{\theta}(\x{} \mid \z{})$ learns the mapping from the latent space to observed space using the samples generated by $p(\z{})$ and model parameters $\theta$. This setup is used to generate new samples not present in the observed dataset. Thus, the aim is to determine the correct setting of the parameters, $\theta$, such that the probability of each observed data, $p_{\theta}(\x{})$, is maximized. The objective function of the DLVM is defined as follows:
\begin{align}
&\max_{\theta} \mathbb{E}_{\x{} \sim p(\x{})} \log{p_{\theta}(\x{})} 
= \max_{\theta} \mathbb{E}_{\x{} \sim p(\x{})} \log{ \int p_{\theta}(\x{} \mid \z{})p(\z{})d\z{}} \nonumber \\
&= \max_{\theta, q} \mathbb{E}_{\x{} \sim p(\x{})} \log \int {\left( p_{\theta}(\x{} \mid \z{})\frac{p(\z{})}{q(\z{})} \right) q(\z{}) d\z{}}\;\;\; \parbox{1.5 in}{\raggedright\scriptsize{Expectation under the proposal distribution}, $q(\z{})$} \label{eq:MLE_proposal_dist} \\
&= \max_{\theta, q} \mathbb{E}_{\x{} \sim p(\x{})} \log \mathbb{E}_{\z{} \sim q(\z{})} \left( p_{\theta}(\x{} \mid \z{})\frac{p(\z{})}{q(\z{})} \right) \nonumber \\
&\text{ by Jensen's inequality, we get } \nonumber \\
&\geq \max_{\theta, q} \mathbb{E}_{\x{} \sim p(\x{})} \mathbb{E}_{\z{} \sim q(\z{})} \log \left( p_{\theta}(\x{} \mid \z{})\frac{p(\z{})}{q(\z{})} \right) \nonumber \\
&= \max_{\theta, q} \mathbb{E}_{\x{} \sim p(\x{})} \mathbb{E}_{\z{} \sim q(\z{})} \left\{ \log \left( p_{\theta}(\x{} \mid \z{}) \right) - \log \left( \frac{q(\z{})}{p(\z{})} \right) \right\} \nonumber \\
&= \max_{\theta, q} \mathbb{E}_{\x{} \sim p(\x{})} \left\{ \mathbb{E}_{\z{} \sim q(\z{})} \log \left( p_{\theta}(\x{} \mid \z{}) \right) - \operatorname{KL} \left( q(\z{}) \lvert\rvert p(\z{}) \right) \right\}.  \label{eq:MLE_ELBO}
\end{align}
The objective function defined in \ref{eq:MLE_ELBO} gives a lower bound on the data log likelihood and is known as the evidence lower bound (ELBO).

Use of $q(\z{} \mid \x{})$ as the proposal distribution in equation \ref{eq:MLE_proposal_dist} gives us the objective function of the VAE \cite{rezende2014stochastic, kingma2014auto}. The choice of the probability distribution for $q$ is a modeling choice, and for VAEs, it is typically a Gaussian distribution \cite{rezende2014stochastic, kingma2014auto}. The VAE uses an inference network (also called a recognition model), $q_{\phi}(\z{} \mid \x{})$, a deep neural network parameterized by $\phi$ that estimates the parameters of the Gaussian distribution for any input $\x{}_{i}$, $\phi: \x{}_{i} \rightarrow (\boldsymbol\mu_{\x{}_{i}}, \boldsymbol\sigma_{\x{}_{i}}^2\Id)$.

Matching the conditional distribution $q_{\phi}(\z{} \mid \x{})$ to $p(\z{})$ in VAEs often fails to match the aggregate posterior in the latent space \cite{rosca2018distribution, ELBO_surgery_2016}. The mismatch leads to, among other things, \emph{holes} or \emph{pockets} in the latent distribution that subsequently affects the quality of the generated samples. Increasing the strength of the regularization term in the objective function of VAEs does not help better match the aggregate posterior to the prior \cite{beta_VAE_ICLR_2017}. Instead, it results in a scenario known as \emph{posterior collapse} \cite{posterior_collapse_ICLR_2019, posterior_collapse_Neurips_2019}, where the conditional distribution $q_{\phi}(\z{} \mid \x{})$ matches to the prior $p(\z{})$ for a subset of the latent dimensions. Such degenerate solutions produce latent encodings that are no longer meaningful, and the decoder tends to ignore $\z{}$ in the reproduced observed data, resulting in poor reconstruction. This phenomenon is related to the identity, $\operatorname{KL} \left( q_{\phi}(\z{} \mid \x{}) \lvert\rvert p(\z{}) \right) = I(\x{};\z{}) + \operatorname{KL} \left( q_{\phi}(\z{}) \lvert\rvert p(\z{}) \right)$, where the $I(\x{};\z{})$ is the mutual information between the observed and latent variables. Thus, increasing the strength of the KL term would lead to better aggregate posterior matching but would lower the mutual information between the latent variables and the data. Several variants are proposed \cite{Factor_VAE_ICML_2018, TC_VAE_Neurips_2019, dai2019_2sVAE} to circumvent these issues encountered in the VAE that emphasizes matching the aggregate posterior to a prior.

\subsection{Aggregate Variational Autoencoder (AVAE)}\label{sec:method_formulation}
Instead of parametric distribution on the conditional probability, as used in VAEs, we propose to represent the aggregate distribution using kernel density estimates (KDE). KDE used to approximate the aggregate posterior distribution is defined as:
\begin{align}\label{eq:kde}
  q(\z{}) &= \frac{1}{m} \sum_{i=1}^m \K{ \frac{||\z{} - \z_{i}^{'}||}{h} }.
\end{align}
Using KDEs, the probability at $\z{}$ for the proposal distribution is estimated using $m$ KDE samples, $\z_{i}^{'}$, and the kernel, $K$, with an associated bandwidth, $h \in \R{}^+$. We use a random subset of the training data, $\mathcal{X{}}^{kde}$, that is shuffled every epoch to produce KDE samples in the latent space, $\z{}_{i}^{'} = \mathbf{E_{\phi}}(\x{}_{i}^{'}), \text{  where } \x{}_{i}^{'} \in \mathcal{X{}}^{kde} \text{ and } \mathbf{E_{\phi}}$ is a deep neural network parameterized by $\phi$, known as the \emph{encoder}. We use a deterministic encoder (ignoring the variances along the latent axes), unlike VAEs. Through multiple empirical evaluations, we show that using a deterministic encoder in the AVAE does not rob it of expressive power compared to a regular VAE or its variants.

The ELBO objective function using KDE-based proposal distribution $q_{\phi}(\z{})$ is defined as follows:
\begin{align}\label{eq:ELBO_agg}
& \max_{\theta, \phi} \mathbb{E}_{\x{} \sim p(\x{})} \left\{ \mathbb{E}_{\z{} \sim q_{\phi}(\z{})} \log \left( p_{\theta}(\x{} \mid \z{}) \right) - \operatorname{KL} \left( q_{\phi}(\z{}) \lvert\rvert p(\z{}) \right) \right\}.
\end{align}
Equation \ref{eq:ELBO_agg} gives us the objective function of the AVAE. In comparison to the proposal distribution in VAEs, KDE-based approximation matches the aggregate posterior, $q_{\phi}(\z{})$, to the prior, $p(\z{})$, \emph{without any modifications} to the ELBO formulation. Compared to the $\beta$-TCVAE \cite{TC_VAE_Neurips_2019}, the AVAE does not have a mutual information (MI) term in its objective function. The absence of the MI in the AVAE also reduces the number of hyperparameters.

The random (data) variable $\x{}$ typically exists in high dimensions, and thus the probability of $p_{\theta}(\x{} \mid \z{})$ is valid only for a small region in the latent space, i.e., $p_{\theta}(\x{} \mid \z{})$ is nonzero for a small region in the latent space. We use $\z{} = \mathbf{E_{\phi}}(\x{})$ as an estimate to maximize $\log p_{\theta}(\x{} \mid \z{})$ in equation \ref{eq:ELBO_agg}. Considering this modeling choice, the objective function of the AVAE then becomes:
\begin{align}\label{eq:AVAE_ELBO}
& \max_{\theta, \phi} \mathbb{E}_{\x{} \sim p(\x{})} \left\{ \log \left( p_{\theta}(\x{} \mid \mathbf{E_{\phi}}(\x{})) \right) - \operatorname{KL} \left( q_{\phi}(\z{}) \lvert\rvert p(\z{}) \right) \right\}.
\end{align}
We use the multivariate Gaussian distribution or Bernoulli distribution as the conditional likelihood distribution, $p_{\theta}(\x{} \mid \z{})$ in \ref{eq:AVAE_ELBO}, depending on the dataset. The parameters of the chosen distribution are estimated using another neural network known as the \emph{decoder}, $\mathbf{D_{\theta}}$, parameterized by $\theta$. The objective function in \ref{eq:AVAE_ELBO} is optimized using the stochastic gradient descent (SGD) that jointly updates the encoder and decoder parameters, $\phi$ and $\theta$, respectively. The first term in the objective function tries to reproduce the input as closely as possible using the corresponding latent statistics (reconstruction loss), while the KL term (matching the aggregate posterior to the prior) regularizes the model parameters.

The objective function of the AVAE is similar to that of WAEs, which have a reconstruction term and a divergence penalty on the aggregate distribution over latent representations. The divergence measure regulates the trade-off between the reconstruction and latent regularization loss. Similar to WAEs, the AVAE has the flexibility in choosing reconstruction cost terms by considering different distributions for $p_{\theta}(\x{} \mid \z{})$. The divergence penalty in the AVAE is the KL divergence, a particular case of the WAE. Nevertheless, the AVAE has provable statistical properties of the latent space, and the proposed method has empirically demonstrated its merit over the WAE under several evaluation metrics discussed in subsequent sections.

\subsubsection{Training:}
The objective function of the AVAE defined in \ref{eq:AVAE_ELBO} has two terms: the reconstruction loss and KL-divergence-matching of the aggregate posterior to the prior. The aggregate posterior, $q_{\phi}(\z{})$, in the AVAE is represented using KDE. A random subset of the training data $\mathcal{X{}}^{kde} \in \mathcal{X{}}^{train}$ forms KDE samples that is shuffled after every epoch. Remaining samples $\mathcal{X{}}^{sgd} = \mathcal{X{}}^{train} - \mathcal{X{}}^{kde}$ are used for optimizing the objective function \ref{eq:AVAE_ELBO} using the SGD that updates the model parameters, $\phi$ and $\theta$. Shuffling KDE samples in $\mathcal{X{}}^{kde}$ in every epoch changes the aggregate posterior, $q_{\phi}(\z{})$, used in the AVAE objective function. However, the evolving aggregate posterior does not impact (adversely) the training of the AVAE, and the loss curves on multiple datasets demonstrate the stable optimization of the AVAE objective function (refer to Figure 1 in the supplementary). Moreover, an update of the $\mathcal{X{}}^{kde}$ and $\mathcal{X{}}^{sgd}$ in every epoch results in better performance of the AVAE (compared with a fixed $\mathcal{X{}}^{kde}$) under different metrics across datasets (refer to Table 1 in the supplementary).

Without any loss of generality, we use the isotropic Gaussian kernel in KDE for this work, which introduces a bandwidth parameter. There are many heuristics for estimating the kernel bandwidth used in KDE, and there is no established solution for unknown distributions. Furthermore, the estimation of KDE bandwidth is particularly challenging in high-dimensional latent spaces (dimensions > 50). We present a bandwidth estimation method in section \ref{sec:AVAE_KDE_bw} that uses the knowledge of the prior distribution, $p(\z{})=\Gauss{\zero}{\Id}$, to estimate KDE bandwidth for a given latent dimension and a given number of KDE samples. The proposed bandwidth estimation technique can even scale to higher dimensional latent spaces, i.e., dimensions > 100.

Several extensions of the VAE \cite{KL_annealing_VAE_ICML_2016, Cycling_KL_annealing_VAE_NAACL_2019,
Control_VAE_ICML_2020, Sigma_VAE_ICML_2021} propose automated ways to determine the hyperparameter $\beta$ that balances the loss terms in the objective function. In a similar vein, we propose a data-driven technique to determine $\beta$ that balances the loss terms in the AVAE objective function. An outline of the training of the AVAE is presented in Algorithm \ref{alg:AVAE_Alg}.

\begin{algorithm}[t]
\caption{: AVAE training using stochastic gradient descent}
\hspace{\algorithmicindent}\textbf{Input:} Data $\mathcal{X}$, Latent dimensions $l$, KDE samples $m$.
\begin{algorithmic}[1]
\STATE Estimate the optimal kernel bandwidth $h_{\rm opt}^{\rm corr}$ given $(l,m)$
\STATE Split $\mathcal{X{}}$ into training, $\mathcal{X{}}^{train}$, and validation data, $\mathcal{X{}}^{val}$
\STATE Random samples for KDE, $\mathcal{X{}}^{kde}$, and SGD samples, $\mathcal{X{}}^{sgd} = \mathcal{X{}}^{train} - \mathcal{X{}}^{kde}$
\STATE Random initialization of the encoder ($\phi$) and decoder ($\theta$) parameters
\STATE Get KDE samples, $\z{}_i^{'} = E_\phi(\x{}_i^{'})$, where $\x{}_i^{'} \in \mathcal{X{}}^{kde}$ for the aggregate posterior, $q_{\phi}$
\STATE Initialize $\beta \gets \frac{1}{n_{val}} \sum_{i=1}^{n_{val}} \| \x{}_{i}^{''} - \hat{\x{}}_{i}^{''} \|_{2}$, where $\x{}_{i}^{''} \in \mathcal{X{}}^{val}$
\FOR {number of epochs}
\FOR {number of minibatch updates}
\STATE Sample a minibatch of size $n_b$ from $\mathcal{X{}}^{sgd}$, $\mathcal{X}_b^{sgd} = \{\x{}_1^{''}, \ldots \x{}_{n_b}^{''}\}$
\STATE Encode samples $\mathcal{Z}_b^{sgd} = \{\z{}^{''}_1, \ldots \z{}^{''}_{n_b}\}$, where $\z{}^{''}_i = E_\phi(\x{}^{''}_i)$
\STATE Update the encoder and decoder parameters, $\phi$ and $\theta$, respectively, by optimizing the objective function in \ref{eq:AVAE_ELBO} using stochastic gradient descent
\STATE Update KDE samples, $\z{}^{'}_i = E_\phi(\x{}^{'}_i)$ using the current state of the encoder
\ENDFOR
\STATE $\beta \gets \frac{1}{n_{val}} \sum_{i=1}^{n_{val}} \| \x{}_{i}^{''} - \hat{\x{}}_{i}^{''} \|_{2}$, where $\x{}_{i}^{''} \in \mathcal{X{}}^{val}$
\STATE New samples for $\mathcal{X{}}^{kde}$ chosen at random and update $\mathcal{X{}}^{sgd} = \mathcal{X{}}^{train} - \mathcal{X{}}^{kde}$
\STATE Produce latent encoding $\z{}^{'}_i = E_\phi(\x{}^{'}_i)$ for the aggregate posterior, $q_{\phi}$, using KDE
\ENDFOR
\end{algorithmic}
\label{alg:AVAE_Alg}
\end{algorithm}

\subsubsection{Estimation of $\beta$:}\label{sec:AVAE_beta}
The objective function of the standard VAE does not introduce a hyperparameter to weigh the loss terms. However, it is a common practice to assign weights to different terms in the objective functions \cite{tolstikhin2017wasserstein, RAE, makhzani2016adversarial} for various reasons, such as stability in optimization and application-specific trade-offs. Likewise, several variants of the VAE \cite{beta_VAE_ICLR_2017, TC_VAE_Neurips_2019} use a hyperparameter, $\beta$, to control the contribution of the loss terms in the objective function. It is often challenging to decide the appropriate value of these hyperparameters for a particular model architecture, dataset, and other related settings for optimization. The widely used strategy under these circumstances is to set the hyperparameter value using cross-validation.

To alleviate these issues, methods proposed in \cite{Control_VAE_ICML_2020, Sigma_VAE_ICML_2021}, among others, have devised automated strategies to determine $\beta$. The method in \cite{Control_VAE_ICML_2020} uses a PI controller that manipulates the value of $\beta$ as the learning progresses. Assuming the decoder predicts the parameter of the multivariate Gaussian distribution, \cite{Sigma_VAE_ICML_2021} presents two approaches to learning the Gaussian variance, $\sigma$ (equivalent to learning $\beta$). In the first approach, an additional parameter is trained with the encoder-decoder parameters to learn the trade-off factor, $\sigma$. In another approach, the maximum likelihood estimate (MLE) determines the variance analytically.

Similar to these approaches, the proposed AVAE optimization sets beta $\beta$ to weight the gradient of the regularization term relative to the reconstruction loss:
\begin{equation}\label{eq:AVAE_beta}
    \beta \leftarrow \frac{1}{n_{val}} \sum_{i=1}^{n_{val}} \| \x{}_{i}^{''} - \hat{\x{}}_{i}^{''} \|_{2}, 
\end{equation}
where $\x{}_{i}^{''}$ is an example in the validation set, $\mathcal{X}^{val} \in \mathcal{X{}}$, and $\hat{\x{}}_{i}^{''}$ is the corresponding reconstructed sample produced by the decoder. Relative to  \cite{Control_VAE_ICML_2020}, the proposed approach is simple yet effective, as demonstrated by the empirical evaluations. The update of $\beta$ during the training of the AVAE on multiple datasets is reported in the supplementary (Figure 1). Moreover, this formulation can be extended to any distribution chosen for the log conditional likelihood, $p(x \mid z)$, rather than being limited to only a Gaussian, as in \cite{Sigma_VAE_ICML_2021}.

\subsection{Properties of the Aggregate Posterior of the AVAE}\label{sec:AVAE_KDE_proof}
Considering the standard normal distribution, $\Gauss{\zero}{\Id}$, as the prior distribution, $p(\z{})$, we analytically derive the expected aggregate posterior distribution of a trained AVAE. For a trained AVAE model, we assume the gradient of the objective function (\ref{eq:AVAE_ELBO}) w.r.t to latent encodings, $\z{}^{''}_i$'s (refer to Algorithm \ref{alg:AVAE_Alg}), is zero. In our analysis, we consider only the KL divergence term in the objective function. Setting the derivative of the $\operatorname{KL} (q_{\phi}(\z{}^{''}) \lvert\rvert p(\z{}^{''}))$ to $0$, we derive the same expression as in equation $5$ of \cite{gens_saha_2022}. Following the steps in \cite{gens_saha_2022}, we prove the aggregate posterior distribution of the AVAE is $\Gauss{\zero}{\Id(1-h^2)}$, in expectation, where $h$ is KDE bandwidth. The proof is also consistent with the known properties of KDEs generally --- KDEs introduce a bias that is characterized by a convolution of the kernel with the underlying distribution.

\subsection{KDE Bandwidth Estimate}\label{sec:AVAE_KDE_bw}
Estimating KDE bandwidth can be challenging, and solutions in the literature are often related to particular applications. Many heuristics are proposed for bandwidth estimation under general circumstances \cite{var_KDE}. However, here, we use the knowledge of the prior distribution, $p(\z{})=\Gauss{\zero}{\Id}$, to our advantage for estimating KDE bandwidth used for modeling the aggregate distribution in the latent space, $q_{\phi}(\z{})$. We devise an objective function such that the empirical aggregate distribution, $q_{\phi}(\z{})$, in the latent space approaches the target distribution, $p(\z{})=\Gauss{\zero}{\Id}$, as the system converges. Thus, we set the kernel bandwidth $h_{\rm{opt}}$  to minimize the KL divergence between the analytical prior distribution and KDE of a finite set of samples from the prior distribution, as follows:
\begin{equation}\label{eq:h_opt}
h_{\rm{opt}} = \min_{h}\operatorname{KL} \left( p(\z{}) \lvert\rvert q_{\phi}(\z{}) \right) = \max_{h} \mathbb{E}_{\z{} \sim p(\z{})} q_{\phi}(\z{}),
\end{equation}
with latent dimension, $l$, and a number of KDE samples, $m$. In this optimization problem, we use samples from the $p(\z{})=\Gauss{\zero}{\Id}$ such that the probability of the samples is maximized w.r.t the aggregate posterior, $q_{\phi}(\z{})$. Table \ref{tab:corr_bandwidth} reports the optimum bandwidth, $h_{\rm{opt}}$, for different scenarios. We use gradient-based optimizers, such as Adam, to learn the single parameter, $h_{\rm{opt}}$ in \ref{eq:h_opt}. We observe in Table \ref{tab:corr_bandwidth} that for higher latent dimensions with limited KDE samples (e.g., starting at $l=40$ with $m=500$), the optimal bandwidth is greater than the standard deviation of the prior distribution, $h_{\rm{opt}} \gt 1.0$. Given the known bias KDE introduces in the AVAE optimization, optimizing the encoder under these conditions would degenerate to samples converging at the origin (posterior collapse).

Given $h_{\rm{opt}}$, we know from section \ref{sec:AVAE_KDE_proof} that the distribution in the latent space of the AVAE converges to $\Gauss{\zero}{\Id(1-h_{\rm{opt}}^2)}$, where $(1-h_{\rm{opt}}^2)$ is bias introduced by KDE. However, we could not consider $\Gauss{\zero}{\Id(1-h_{\rm{opt}}^2)}$ as the target distribution, $p(\z{})$, for optimization of the objective function in \ref{eq:h_opt}, as $h_{\rm{opt}}$ is unknown. We hypothesize that this is one of the reasons for the optimal bandwidth $h_{\rm{opt}}$ to be greater than the standard deviation of the prior distribution in bigger latent spaces (Table \ref{tab:corr_bandwidth}). Thus, we must factor in the bias, $(1 - h_{\rm{opt}}^{2})$, introduced by KDE to estimate the bandwidth. To this end, we propose to use a scaled version of the target distribution, $\Gauss{\zero}{\alpha^2\Id}$, for optimization of the objective function in \ref{eq:h_opt}, where the scaling factor $\alpha$ is unknown. We need to estimate the optimum bandwidth for $\Gauss{\zero}{\alpha^2\Id}$. Given $h_{\rm{opt}}$ as the optimum bandwidth for $\Gauss{\zero}{\Id}$, the estimated bandwidth for $\Gauss{\zero}{\alpha^2\Id}$ is $\alpha h_{\rm{opt}}$ by linear property of the Gaussian distribution. Moreover, we know (from section \ref{sec:AVAE_KDE_proof}) that with $\alpha h_{\rm{opt}}$ as KDE bandwidth, the latent distribution of the AVAE would have a bias, $1-(\alpha h_{\rm{opt}})^{2}$, at convergence. We use this property to solve for the scaling factor, $\alpha$, where we set the variance equal to the bias, $\alpha^{2} = 1 - (\alpha h_{\rm{opt}})^{2}$, to get the scaling that accounts for both the ideal optimal bandwidth and the bias:
\begin{equation}
    \alpha^{2} = \frac{1}{1 + h_{\rm{opt}}^{2}}.
\end{equation}
This simple but elegant strategy of handling the bias in KDE addresses the challenge of estimating KDE bandwidth in high-dimensional latent spaces. Notice that because $0 \le \alpha \le 1.0$, we avoid mode collapse because the system only degenerates ($h_{\rm{opt}} \rightarrow \infty$) as the number of samples goes to zero or the dimensionality goes to infinity.

\begin{table}[t]
    \centering
    \caption{Optimal bandwidths, $h_{\rm opt}$ (estimated using the objective function defined in \ref{eq:h_opt}) and corresponding bias-corrected estimations $h_{\rm{opt}}^{\rm{corr}}$ ($h_{\rm opt}$ scaled by the factor $\alpha$) for a given latent dimension ($l$) and number of KDE samples ($m$). The estimated bandwidth increases with increasing dimensions (vertical) and decreases with increasing sample size (horizontal). For higher latent dimensions with limited KDE samples (e.g., starting at $l=40$ with $m=500$), $h_{\rm{opt}} \gt 1.0$. However, the bias-corrected bandwidth $h_{\rm{opt}}^{\rm{corr}} \lt 1.0$.}\label{tab:corr_bandwidth}
    {%
    \begin{tabular}{|c c || cc | c || cc cc | c || cc cc | c || cc cc | c || cc cc | c |}
    \hline
    &&\multicolumn{3}{c||}{$m=500$}&&&\multicolumn{3}{c||}{$m=1000$}&&&\multicolumn{3}{c||}{$m=2000$}&&&\multicolumn{3}{c||}{$m=5000$}&&&\multicolumn{3}{c|}{$m=10000$}\\ 
    \cline{3-25}
    $l$&&$h_{\rm opt}$&&$h_{\rm opt}^{\rm corr}$ &&&$h_{\rm opt}$&&$h_{\rm opt}^{\rm corr}$ &&&$h_{\rm opt}$&&$h_{\rm opt}^{\rm corr}$ &&&$h_{\rm opt}$&&$h_{\rm opt}^{\rm corr}$ &&&$h_{\rm opt}$&&$h_{\rm opt}^{\rm corr}$\\\hline\hline
    10&&$0.74$&&$0.60$ &&&$0.70$&&$0.58$ &&&$0.67$&&$0.56$ &&&$0.63$&&$0.53$ &&&$0.60$&&$0.51$ \\ \hline
    20&&$0.89$&&$0.67$ &&&$0.86$&&$0.65$ &&&$0.84$&&$0.64$ &&&$0.80$&&$0.62$ &&&$0.78$&&$0.61$ \\ \hline
    40&&$> 1.0$&&0.72 &&&$> 1.0$&&0.71 &&&$0.98$&&$0.70$ &&&$0.95$&&$0.69$ &&&$0.93$&&$0.68$ \\ \hline
    50&&$ > 1.0$&&$0.73$ &&&$ > 1.0$&&$0.72$ &&&$ > 1.0$&&$0.71$ &&&$0.99$&&$0.70$ &&&$0.98$&&$0.70$ \\ \hline
    70 &&$ > 1.0$&&$0.74$ &&& $> 1.0$&&$0.74$ &&& $ > 1.0 $&&$0.73$ &&& $ > 1.0$&&$0.73$ &&& $ > 1.0$&&$0.72$ \\ \hline
    100 && $ > 1.0$&&$0.76$&&&$> 1.0$&&$0.75$ &&&$ > 1.0$&&$0.75$ &&&$ > 1.0$&&$0.74$ &&&$ > 1.0$&&$0.74$ \\ \hline
    \end{tabular}
    }
\end{table}

With the {\em bias scaling factor}, $\alpha$, we get estimates of the \emph{bias-corrected} KDE bandwidth ($h_{\rm{opt}}^{\rm{corr}}=\alpha*h_{\rm{opt}}$) reported in Table \ref{tab:corr_bandwidth}, which are the scaled versions of the optimum bandwidth $h_{\rm{opt}}$ (equation \ref{eq:h_opt}). In the revised estimate, the optimal bandwidth is less than the standard deviation of the prior distribution, $h_{\rm{opt}}^{\rm{corr}} \lt 1.0$, for all dimensions in Table \ref{tab:corr_bandwidth}, as expected. The bias-corrected bandwidth encourages the use of KDEs in bigger latent spaces (e.g., dimensions $\ge 50$) that makes the AVAE appropriate for modeling complex datasets (a limitation in the previous KDE-based aggregate matching \cite{gens_saha_2022}).

\section{Experiments}\label{sec:exp_sec}
\subsection{Experimental Setup}\label{sec:exp_setuo}
\textbf{Benchmark Methods:} 
In comparisons, we consider the conventional VAE \cite{kingma2014auto} and other variations of VAE that modify the original formulation in an attempt to match the aggregate posterior to the prior \cite{Factor_VAE_ICML_2018, TC_VAE_Neurips_2019}. Among others, $\beta$-TCVAE \cite{TC_VAE_Neurips_2019} is the closest to the AVAE formulation, as the objective function does not introduce any additional, ad-hoc loss terms. The RAE \cite{RAE} is chosen as one of the baseline models due to its performance on multiple benchmark datasets. Other maximum likelihood-based models such as the AAE \cite{makhzani2016adversarial} and WAE \cite{tolstikhin2017wasserstein} match aggregate posterior in the latent space of a deterministic autoencoder. The AAE \cite{makhzani2016adversarial} implicitly matches aggregate distributions using a discriminator in the latent space. We use the WAE-MMD (with IMQ kernel) in our analysis due to the stability in training. We study the VAE \cite{kingma2014auto}, $\beta$-TCVAE \cite{TC_VAE_Neurips_2019}, RAE \cite{RAE}, AAE\cite{makhzani2016adversarial}, and WAE-MMD \cite{tolstikhin2017wasserstein} as competing methods to the AVAE.

\textbf{Evaluation Metrics:} Ideally, the evaluation of a DLVM should include a comparison of the model's data distribution and that of the true data. Of course, this is infeasible because true data distribution is unknown. Many methods use the quality of the samples produced by the models in the observed space as a proxy for the actual distribution. In this work, we use the Fréchet Inception Distance (FID) \cite{IS_FID_1} to quantify the quality of the samples. In addition, we evaluate the data distributions learned by different models using the precision-recall metric \cite{precision_recall_distributions}, where the precision evaluates the quality of the generated samples, and the recall assesses whether the model data distribution captures the variations present in the original but unknown data distribution. Besides the attributes of the model data distribution, we evaluate the properties of the latent representations of the competing methods. In particular, we are interested in the presence of holes in the latent distribution, and we use \emph{entropy} of the aggregate posterior distribution as an indicator of holes/clusters. We train each method $5$ times on a dataset for all empirical evaluations, initialized differently in every run.

\textbf{Datasets:} We use several popular benchmark datasets, MNIST \cite{MNIST}, CelebA \cite{CelebA}, and CIFAR10 \cite{CIFAR10} for empirical evaluation of different methods. To address the dataset's complexity, the size of the latent space, neural network architectures, model-specific hyperparameters, and other optimization parameters are altered accordingly. Details of the neural network architectures and other parameter settings for all the benchmark datasets used by the competing methods are reported in sections 2.1 and 2.2 in the supplementary material.

\subsection{Results}\label{sec:results}
\subsubsection{Evaluation of the Model Data Distribution}\label{sec:FID_score} We quantitatively evaluate the generated samples in this experiment using the FID scores \cite{IS_FID_1} on multiple benchmark datasets. A lower FID score indicates better matching of the data distributions. Besides the FID metric, we evaluate the diversity and quality of the generated samples using the precision-recall metric \cite{precision_recall_distributions}. A higher precision indicates good quality of the generated samples, and a higher recall suggests that the model data distribution covers the modes present in the true data distribution. Except for the RAE, all the methods considered in this experiment use $\Gauss{\zero}{\Id}$ as the prior distribution. For the RAE, we approximate the distribution in the latent space by the Gaussian distribution. Parameters of the Gaussian distribution derived from the latent representations are used to generate new data samples. We know that the latent distribution for the AVAE convergences to $\Gauss{\zero}{\Id(1-h^2)}$, where h is KDE bandwidth (section \ref{sec:AVAE_KDE_proof}). Therefore, we use samples drawn from the distribution, $\Gauss{\zero}{\Id(1-h^2)}$, to evaluate the generative capability of AVAEs. For a fair comparison, we have used the hyperparameter settings suggested by the author or recommended in the literature. 

\begin{table*}[t]
    \centering
    \caption{FID \cite{IS_FID_1}, and precision-recall \cite{precision_recall_distributions} scores of competing methods. The \textbf{best} score is in \textbf{bold}, and the \underline{second best} score is \underline{{underlined}}.} \label{tab:FID_Score}
    \resizebox{\textwidth}{!}
    {%
    \begin{tabular}{c c c c c c c c c c c c c}
    \hline
    \multirow{2}{*}{\parbox{1.5cm}{\centering Methods}}&&\multicolumn{3}{c}{MNIST\;\;$(l=16)$}&&\multicolumn{3}{c}{CelebA\;\;$(l=64)$}&&\multicolumn{3}{c}{CIFAR10\;\;$(l=128)$}\\ \cline{3-5}\cline{7-9}\cline{11-13}
    &&\textsc{FID} $\downarrow$&\textsc{Precision}$\uparrow$&\textsc{Recall}$\uparrow$&&\textsc{FID}$\downarrow$&\textsc{Precision}$\uparrow$&\textsc{Recall}$\uparrow$&&\textsc{FID}$\downarrow$&\textsc{Precision}$\uparrow$&\textsc{Recall}$\uparrow$\\ \hline
    VAE && $28.78 \pm 0.48$ & $\underline{0.88 \pm 0.04}$ & $\underline{0.97 \pm 0.00}$
    && $49.89 \pm 0.57$ & $0.79 \pm 0.03$ & $0.75 \pm 0.03$
    && $147.74 \pm 0.81$ & $0.50 \pm 0.03$ & $0.47 \pm 0.02$ \\
    $\beta$-TCVAE && $50.62 \pm 1.19$ & $0.82 \pm 0.03$ & $0.95 \pm 0.01$
    && $50.14 \pm 0.78$ & $0.78 \pm 0.02$ & $0.70 \pm 0.05$
    && $180.94 \pm 1.16$ & $0.30 \pm 0.01$ & $0.41 \pm 0.03$ \\
    RAE && $\underline{18.79 \pm 0.31}$ & $0.87 \pm 0.01$ &$0.95 \pm 0.02$ 
    && $\underline{48.81 \pm 1.02}$ & $0.81 \pm 0.02$ & $\underline{0.77 \pm 0.04}$
    && $\underline{94.34 \pm 1.58}$ & $\mathbf{0.74 \pm 0.02}$ & ${0.47 \pm 0.04}$ \\
    AAE && $19.51 \pm 1.77$ & $0.85 \pm 0.03$ &$0.96 \pm 0.01$ 
    && $49.32 \pm 0.25$ & $\underline{0.86 \pm 0.01}$ & $0.75 \pm 0.03$
    && $100.00 \pm 1.40$ & $0.71 \pm 0.03$ & $\underline{0.56 \pm 0.04}$  \\
    WAE && $25.42 \pm 1.19$ & $\mathbf{0.92 \pm 0.03}$ & $0.92 \pm 0.01$
    && $72.01 \pm 2.26$ & $0.64 \pm 0.05$ & $0.75 \pm 0.02$
    && $140.49 \pm 0.64$ & $0.42 \pm 0.01$ & $0.31 \pm 0.05$ \\
    AVAE && $\mathbf{13.27 \pm 0.34}$ &$\mathbf{0.92 \pm 0.02}$ &$\mathbf{0.98 \pm 0.00}$
    && $\mathbf{46.0 \pm 0.42}$ & $\mathbf{0.88 \pm 0.02}$ & $\mathbf{0.85 \pm 0.02}$ 
    && $\mathbf{90.93 \pm 6.65}$ & $\underline{0.72 \pm 0.05}$ & $\mathbf{0.67 \pm 0.04}$ \\ [1.5ex] \hline
    \end{tabular}}
\end{table*}

The FID and precision-recall scores are reported in Table \ref{tab:FID_Score}. The VAE does reasonably well for the MNIST and CelebA datasets. However, its performance drops significantly for the complex CIFAR10 dataset. Despite the importance given to match the aggregate posterior in the $\beta$-TCVAE ($\beta=2$ for comparable reconstruction loss), it fails to address the shortcomings of the VAE. Furthermore, the performance of the $\beta$-TCVAE is poorer than the regular VAE. These results manifest the limitations in the formulation of the VAE (objective function and modeling assumptions) to model the data distributions. Other DLVMs (AAE, WAE, and AVAE) \emph{matching the aggregate posterior} to the prior using a deterministic autoencoder do better than VAEs, in general. The AAE (aka WAE-GAN) closely follows the best performing methods under different evaluation metrics. We hypothesize that the kernel-based method used in WAE (WAE-MMD) to evaluate the mismatch between distributions is possibly leading to poor performance (justified by low entropy scores in Table \ref{tab:entropy_latent_space}), as the reconstruction error is comparable to all other methods (refer to the MSE per pixel in Table 5 of the supplementary material). The performance of the WAE gets worse for the CIFAR10 dataset using high-dimensional latent space $l=128$. The performance of the RAE is promising across all datasets under different evaluation scenarios. The generative capability of the AVAE is the best among all the considered methods for all the benchmark datasets under different evaluation metrics studied in this work, except for the precision on the CIFAR10 dataset (the second best). It is important for any generative model to capture the modes present in a dataset, indicated by high recall scores. The AVAE consistently outperforms other methods under the recall metric, resulting in the best FID scores under all evaluation scenarios.

We investigate the poor performance of the VAE and $\beta$-TCVAE on the MNIST and CIFAR10 datasets. Other than the CelebA dataset, we observe the reconstruction loss of the VAE and $\beta$-TCVAE to be relatively higher than other methods (refer to Table 5 in the supplementary material). On further analysis, we discovered that both the VAE and $\beta$-TCVAE suffer from the posterior collapse when trained on the MNIST and CIFAR10 datasets (refer to section 2.3 in the supplementary). For the MNIST dataset, $4$ and $7$ (out of $16$) latent dimensions collapsed for the VAE and $\beta$-TCVAE, respectively. Collapsed dimensions reduce the bottleneck capacity of a DLVM, resulting in higher reconstruction loss. The posterior collapse subsequently impairs the VAE and $\beta$-TCVAE to model the data distributions, leading to the worst FID scores for the $\beta$-TCVAE on the MNIST and CIFAR10 datasets, followed by the VAE.

\begin{table*}[t]
    \centering
    \caption{Mean entropy of the $q_{\phi}^{\rm{w}}(\z{})$ produced by competing methods on the benchmark datasets. The \textbf{best} score is in \textbf{bold}, and the \underline{second best} score is \underline{{underlined}}. The entropy of the standard normal distribution is used as the ground truth.} \label{tab:entropy_latent_space}
    {%
    \begin{tabular}{|c||c||c||c|}\hline
    Method & MNIST $(l=16) \uparrow$ & CelebA $(l=64) \uparrow$ & CIFAR10  $(l=128) \uparrow$ \\ [0.5ex] 
    \hline\hline
    VAE             & $4.71 \pm 0.14$ & $28.88 \pm 0.03$ & $29.56 \pm 0.39$ \\
    $\beta$-TCVAE   & $ 3.44 \pm 0.39 $ & $ 28.42 \pm 0.05 $ & $ 15.02 \pm 0.69 $ \\
    RAE             & $4.89 \pm 0.06$ & $27.08 \pm 0.09$ & $49.92 \pm 0.31$ \\
    AAE             & $\underline{5.81 \pm 0.18}$ & $\underline{30.49 \pm 0.03}$ & $\underline{54.21 \pm 0.36}$ \\
    WAE             & $4.67 \pm 0.09$ & $28.32 \pm 0.05$ & $48.10 \pm 0.45$ \\
    AVAE            & $\mathbf{7.56 \pm 0.10}$ & $\mathbf{30.96 \pm 0.02}$ & $\mathbf{55.64 \pm 0.00}$ \\
    Standard Normal & $8.00 $ & $32.00 $ & $64.00 $\\ [1.5ex] \hline
    \end{tabular}}
\end{table*}

\subsubsection{Entropy of the Aggregate Posterior Distribution} In this experiment, we evaluate deviations of the resultant aggregate distribution, $q_{\phi}(\z{})$, beyond the second moment (other than the mean and covariance), as we would expect from holes or clusters in the distribution. For this, we use the entropy of the aggregate posterior distribution to quantify how close it is to Gaussian, {\em after whitening the distribution}, $q_{\phi}^{\rm{w}}(\z{})$, to remove the effects of the second moment mismatch. Because the Gaussian distribution has the maximum entropy (for a given mean and covariance), we use the entropy of the whitened data. Entropy is defined as
\begin{align}\label{eq:entropy_def}
H(\z{}) & = \mathbb{E}_{\z{} \sim q_{\phi}^{\rm{w}}(\z{})} \left\{ - \log \left( q_{\phi}^{\rm{w}}(\z{}) \right) \right\} \approx \frac{1}{m} \sum_{j} \frac{1}{m-1} \sum_{i \ne j} \K{ \frac{||\z_{j} - \z_{i}^{'}||}{h} },
\end{align}
where $q_{\phi}^{\rm{w}}(\z{})$ is the aggregate posterior distribution over the whitened data. We use KDE (defined in \ref{eq:kde}) for estimating the density $q_{\phi}^{\rm{w}}(\z{})$ for all methods because it can, in principle model the deviations we are seeking to evaluate. The bandwidth $h$ required in KDE for the latent dimensions $l=\{16, 64, 128\}$ (for different datasets) and KDE samples $m=10K$ is derived using the strategy defined in section \ref{sec:AVAE_KDE_bw}. The entropy computation uses the held-out set of the datasets studied in this work. The entropy of the standard normal distribution  (leaving out the constants) derived analytically serves as the ground truth.

\begin{figure*}[t]
    \centering
    \includegraphics[width=1.0\textwidth]{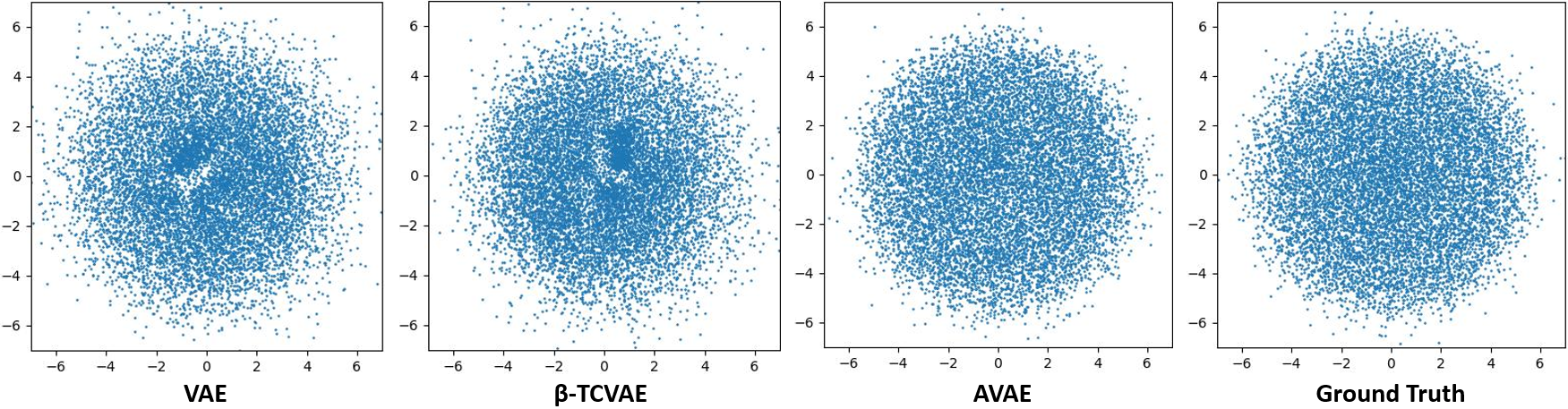}
    \caption{The metric multidimensional scaling (mMDS) \cite{metric_MDS} plot in 2D of the latent representations ($\mathcal{Z} \in \R{}^{16}$) produced by the VAE \cite{kingma2014auto}, $\beta$-TCVAE \cite{TC_VAE_Neurips_2019} and the AVAE (proposed method) on the MNIST dataset \cite{MNIST}. Samples from the target distribution, $\Gauss{\zero}{\Id}$, are used as the ground truth. The regions of low probability and unwanted aggregation of data points in different parts of the latent space of the VAE and $\beta$-TCVAE clearly show the mismatch with the ground truth. The AVAE closely matches the target distribution corroborated by empirical evaluations.}
    \label{fig:Holes_VAE}
\end{figure*}

From the results reported in Table \ref{tab:entropy_latent_space}, we observe that the entropy scores of the VAE and $\beta$-TCVAE are far off from the ground truth for the MNIST and CIFAR10 datasets even using the whitened latent representations. Low entropy scores of the VAE and $\beta$-TCVAE can be attributed to the formation of clusters as observed in Figure \ref{fig:Holes_VAE} (refer to section 2.3 in the supplementary for more results). Besides the posterior collapse, the entropy scores offer another perspective to explain the high FID scores of the generated samples produced by the VAE and $\beta$-TCVAE for the MNIST and CIFAR10 datasets. Poor FID scores of the WAE can be related to the low entropy values across datasets. The low entropy scores of the RAE are not surprising because it does not attempt to match any prior distribution in the latent space. However, the regularization approach in the RAE is more effective than the VAE. The AAE has entropy scores comparable to the AVAE, and it also helps us comprehend the consistent FID scores of the generated samples. The best entropy score of the AVAE for all the datasets indicates the close matching of the aggregate posterior to the prior, as shown in Figure \ref{fig:Holes_VAE}, where we do not observe clustering of the latent representations. 

\begin{table}[t]
    \centering
    \caption{Comparison of the performance of the AVAE with \textbf{different} number of KDE samples under multiple metrics for the MNIST and CAIFAR10 datasets.
    } \label{tab:KDE_Samples}
    \resizebox{\textwidth}{!}
    {%
    \begin{tabular}{c c c c c c c c c c c c c}
    \hline
    \multirow{2}{*}{\parbox{1.5cm}{\centering KDE samples}} &&\multicolumn{4}{c}{MNIST\;\;$(l=16)$}&&&\multicolumn{4}{c}{CIFAR10\;\;$(l=128)$}\\ \cline{3-6} \cline{9-12}
    &&\textsc{FID} $\downarrow$&\textsc{Precision}$\uparrow$&\textsc{Recall}$\uparrow$&\textsc{MSE}$\downarrow$
    &&& \textsc{FID}$\downarrow$&\textsc{Precision}$\uparrow$&\textsc{Recall}$\uparrow$&\textsc{MSE}$\downarrow$\\ \hline
    1000 && 
    $14.26 \pm 0.23$ &$0.89 \pm 0.04$ &$0.97 \pm 0.00$ &$0.0033 \pm 0.0002$
    &&& $95.07 \pm 5.52$ & $0.75 \pm 0.05$ & $0.66 \pm 0.03$ &$0.0083 \pm 0.0032$ \\
    2000 && 
    $13.09 \pm 0.73$ &$0.88 \pm 0.03$ &$0.97 \pm 0.01$ &$0.0034 \pm 0.0002$
    &&& $99.88 \pm 15.32$ & $0.70 \pm 0.11$ & $0.67 \pm 0.02$ &$0.0064 \pm 0.0002$ \\
    5000 && 
    $13.64 \pm 1.24$ &$0.89 \pm 0.02$ &$0.98 \pm 0.01$ &$0.0048 \pm 0.0010$
    &&& $103.89 \pm 5.21$ & $0.59 \pm 0.04$ & $0.68 \pm 0.04$ &$0.0065 \pm 0.0003$ \\
    10000 && 
    $13.27 \pm 0.34$ &$0.92 \pm 0.02$ &$0.98 \pm 0.00$ &$0.0041 \pm 0.0004$
    &&& $90.93 \pm 6.65$ & $0.72 \pm 0.05$ & $0.67 \pm 0.04$ &$0.0062 \pm 0.0002$ \\
    [1.5ex] \hline
    \end{tabular}
    }
\end{table}

\subsubsection{Ablation Study} In this experiment, we study the effect of the number of KDE samples on the performance of the AVAE under different evaluation metrics. The number of KDE samples used in the ablation study is $m=1K, 2K, 5K$, and $10K$ for the MNIST and CIFAR10 datasets. We report the FID, precision-recall scores, and the reconstruction loss, i.e., the mean squared error (MSE) per pixel in Table \ref{tab:KDE_Samples}. The AVAE produces comparable results with \emph{a very few} KDE samples, $m=1000$, even in high-dimensional latent space ($l=128$ for the CIFAR-10 dataset). The stable optimization of the AVAE objective function with fewer KDE samples, such as $m=\{1K, 2K\}$ for the MNIST and CIFAR10 datasets, corroborates the accuracy and robustness of the proposed KDE bandwidth estimation technique. Overall, the performance of the AVAE under multiple metrics is slightly better with higher KDE samples. Therefore, we use $m=10K, 20K,$ and $10K$ for the MNIST, CelebA, and CIFAR10 datasets for all the evaluations reported in the paper.

\section{Conclusion}
We propose a novel algorithm, the aggregate VAE (AVAE), based on the framework of the VAE to match the aggregate posterior distribution to the prior using KDE. Using the known properties of the prior distribution, we devised a method to estimate KDE bandwidth in high-dimensional latent spaces (dimensions $\gt 100$) that allows the modeling of complex datasets using the AVAE. The dynamic adjustment of the scaling factor, $\beta$, using the validation data avoids the hyperparameter tuning using cross-validation. The training of the AVAE does not suffer from the posterior collapse, as in VAEs and $\beta$-TCVAEs, and we avoid such failures without the modification of the ELBO formulation \cite{posterior_collapse_delta_VAE_ICLR_2019, resample_prior_VAE, Flex_VAE_Prior_2018} and use of any complex training schedules \cite{KL_annealing_VAE_ICML_2016, Cycling_KL_annealing_VAE_NAACL_2019, Train_VAE_Lag_ICLR_2019}. We demonstrate the efficacy of the proposed method on multiple datasets, and the AVAE consistently outperforms the competing methods under different evaluation metrics. Close matching of the aggregate latent distribution to the prior with comparable reconstruction loss resulted in the best FID, precision, and recall scores for the AVAE. High entropy scores for the AVAE indicate that the latent representations are close to Gaussian and have a lower chance of encountering holes/clusters in the distribution. Through extensive empirical evaluation, we demonstrate the effectiveness of KDE in matching distributions in high-dimensional latent spaces compared to other methods, such as the kernel-based method used in the WAE-MMD and the discriminator in the AAE. In the AVAE, the cardinal latent axes do not represent the generative factors, unlike the regular VAE, due to matching the aggregate posterior to isotropic Gaussian, invariant to rotation. We plan to study this issue and devise a statistical method to identify the latent explanatory vectors.

\end{document}


\title{Supplementary: Matching Aggregate Posteriors in the Variational Autoencoder}
\titlerunning{Supplementary: AVAE}

\author{Surojit Saha\inst{1}\orcidID{0000-0003-0198-0189} \and
Sarang Joshi\inst{1} \and
Ross Whitaker\inst{1}}

\authorrunning{S. Saha et al.}

\institute{Scientific Computing and Imaging Institute, The University of Utah, USA.\\
\email{surojit.saha@utah.edu; sjoshi@sci.utah.edu; whitaker@cs.utah.edu}}

\maketitle              

\section{Training of the AVAE}
\begin{figure*}[h]
    \centering
    \includegraphics[width=1.0\textwidth]{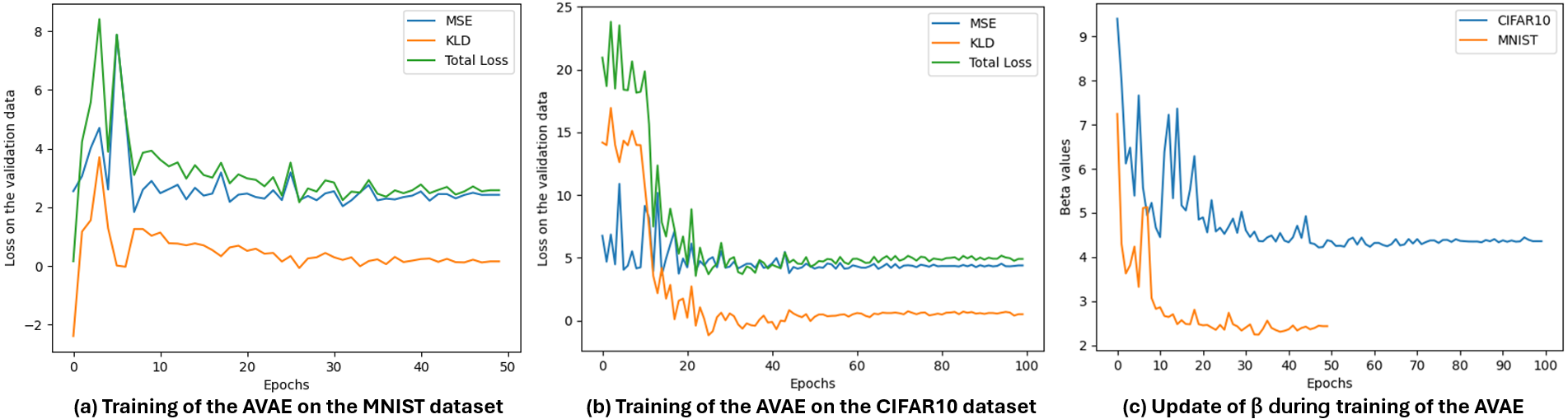}
    \caption{Training of the AVAE with shuffling of KDE samples after every epoch on the (a) MNIST (trained for $50$ epochs) and the (b) CIFAR10 (trained for $100$ epochs) datasets. The loss curves on the validation data indicate stable optimization of the AVAE objective function on multiple datasets. In addition, we show (c) the update of the $\beta$ used in the AVAE objective function on the MNIST and CIFAR10 datasets.}
    \label{fig:AVAE_Training}
\end{figure*}

The aggregate posterior, $q_{\phi}(\z{})$, is estimated using a random subset of the training data $\mathcal{X{}}^{kde} \in \mathcal{X{}}^{train}$ that is shuffled after every epoch (refer to the AVAE algorithm). Remaining samples $\mathcal{X{}}^{sgd} = \mathcal{X{}}^{train} - \mathcal{X{}}^{kde}$ are used for optimizing the AVAE objective function. Shuffling of KDE samples in $\mathcal{X{}}^{kde}$ in every epoch changes the aggregate posterior, $q_{\phi}(\z{})$, used in the AVAE objective function. However, the revised estimate of the aggregate posterior in every epoch does not \emph{affect/hurt} the optimization of the AVAE as shown in Figure \ref{fig:AVAE_Training}. Figure \ref{fig:AVAE_Training} illustrates the stable optimization of the AVAE over multiple datasets (the MNIST and CIFAR10 datasets) along with the update of the $\beta$ values used in the AVAE objective function. 

Furthermore, we empirically demonstrate the advantage of the shuffling strategy in achieving better or comparable performance (under a few scenarios) over multiple datasets. From the results reported in Table \ref{tab:Fixed_vs_Shuffling}, we observe an overall improvement in the performance of the AVAE with the shuffling of KDE samples.

\begin{table}[t]
    \centering
    \caption{Comparison of the performance of the AVAE with \textbf{Fixed} or \textbf{Shuffled} KDE samples under different metrics for the MNIST and CAIFAR10 datasets. We use $5$ different seeds to train models on different datasets and report the average metric score and standard deviation. The \textbf{best} score is in \textbf{bold}.} \label{tab:Fixed_vs_Shuffling}
    \resizebox{\textwidth}{!}
    {%
    \begin{tabular}{c c c c c c c c c c c c}
    \hline
    &&\multicolumn{4}{c}{MNIST\;\;$(l=16)$}&&\multicolumn{4}{c}{CIFAR10\;\;$(l=128)$}\\ \cline{3-6} \cline{8-11}
    &&\textsc{FID} $\downarrow$&\textsc{Precision}$\uparrow$&\textsc{Recall}$\uparrow$&\textsc{MSE}$\downarrow$&&\textsc{FID}$\downarrow$&\textsc{Precision}$\uparrow$&\textsc{Recall}$\uparrow$&\textsc{MSE}$\downarrow$\\ \hline
    Fixed && 
    ${15.00 \pm 2.16}$ &${0.88 \pm 0.03}$ &${0.97 \pm 0.01}$ &$0.0047 \pm 0.0015$
    && ${110.44 \pm 6.85}$ & ${0.64 \pm 0.07}$ & $\mathbf{0.74 \pm 0.04}$ &$0.0068 \pm 0.0002$ \\
    Shuffled && 
    $\mathbf{13.27 \pm 0.34}$ &$\mathbf{0.92 \pm 0.02}$ &$\mathbf{0.98 \pm 0.00}$ &$\mathbf{0.0041 \pm 0.0004}$
    && $\mathbf{90.93 \pm 6.65}$ & $\mathbf{0.72 \pm 0.05}$ & ${0.67 \pm 0.04}$ &$\mathbf{0.0062 \pm 0.0002}$\\ \hline
    \end{tabular}
    }
\end{table}

In another analysis, we estimate the average computation time (in \textit{secs}) taken per epoch to train the AVAE with a different number of KDE samples, $m=1K, 2K, 5K$, and $10K$ for the MNIST and CAIFAR10 datasets using NVIDIA TITAN V ($12$GB). We report the results in Table \ref{tab:AVAE_Time}, where we consider the computation time of the VAE \cite{kingma2014auto} as the baseline for comparison. We observe that the time taken by the AVAE increases with the increase in the number of KDE samples. The time taken by the AVAE is comparable to the VAE for the CIFAR10 dataset, even using $10K$ KDE samples. The computation time of the AVAE using $10K$ KDE samples for the MNIST dataset is significantly higher than the VAE. However, from the results reported in Table 4 of the main paper, we know that there is a marginal improvement in the performance of the AVAE with the increase in the number of KDE samples for the MNIST dataset. Therefore, we can resort to \emph{only} $1K$ KDE samples for the MNIST dataset for faster training.

\begin{table}[t]
    \centering
    \caption{The average computation time (in \textit{secs}) taken per epoch to train the AVAE with different number of KDE samples, $m=1K, 2K, 5K$, and $10K$ for the MNIST and CAIFAR10 datasets using NVIDIA TITAN V ($12$GB). We considered the regular VAE as the baseline.
    } \label{tab:AVAE_Time}
    {%
    \begin{tabular}{c c c c c c c c c c c}
    \hline
    \multirow{2}{*}{\parbox{1.5cm}{\centering Datasets}}
    &&\multirow{2}{*}{\parbox{1.5cm}{\centering VAE\cite{kingma2014auto}}}
    &\multicolumn{8}{c}{AVAE}\\ \cline{4-11}
    &&&&\textsc{$m=1000$} 
    &&\textsc{$m=2000$}
    &&\textsc{$m=5000$} 
    &&\textsc{$m=10000$}\\ \hline
    \textsc{MNIST} 
    &&$7.03 \pm 0.14$ 
    &&$13.43 \pm 0.16$
    &&$14.81 \pm 0.28$
    &&$21.10 \pm 0.67$ 
    &&$27.58 \pm 0.66$ \\
    \textsc{CIFAR10} 
    &&$41.76 \pm 1.16$
    &&$41.67 \pm 0.36$
    &&$42.96 \pm 0.72$
    &&$47.10 \pm 0.37$ 
    &&$53.38 \pm 0.33$ \\
    [1.1ex] \hline
    \end{tabular}
    }
\end{table}

\section{Experiments and Results}
\subsection{Implementation Details}\label{supp:impl_details} For a given dataset, we use the same latent dimension, encoder-decoder architecture, and optimization strategies (such as the learning rate, learning rate scheduler, epochs, and batch size) for all the competing methods to ensure a fair comparison. We train the VAE and $\beta$-TCVAE for more epochs on the MNIST dataset for convergence. The latent dimensions used in the MNIST, CelebA, and CIFAR10 datasets are $l=16, 64,$ and $128$, respectively \cite{tolstikhin2017wasserstein, RAE}. KDE samples used in the AVAE are $m=10K, 20K,$ and $10K$ for the MNIST, CelebA, and CIFAR10 datasets. The aggregate posterior used in the AVAE, $q_{\phi}$, is derived using KDE samples, $\mathcal{X{}}^{kde}$, which are updated in every epoch (refer to the AVAE algorithm). 

\begin{table}[h]
\centering
\caption{Optimization settings for different methods.} \label{tab:hyper-parameters}
{%
\begin{tabular}{c c c c c c c c c}
\hline
Method&&Parameters&&{MNIST}&&{CelebA}&&{CIFAR10}\\ \hline
\textsc{$\beta$-TCVAE} && $\beta$: && $2$ && $2$ && $2$\\
\textsc{WAE} && \textsc{Recons-scalar}: && $0.05$ && $0.05$ && $0.05$\\
\textsc{WAE} && $\beta$: && $10$ && $100$ && $100$\\
\textsc{AAE} && \textsc{Recons-scalar}: && $0.05$ && $0.05$ && $0.05$\\
\textsc{AAE} && $\beta$: && $1$ && $0.1$ && $1$\\
\textsc{RAE} && $\beta$: && $1e-04$ && $1e-04$ && $1e-03$\\
\textsc{RAE} && \textsc{Dec-L}$2$-\textsc{reg}: && $1e-07$ && $1e-07$ && $1e-06$\\
\hline
\end{tabular}
}
\end{table}

The objective function of several methods studied in this work has hyper-parameters associated with the regularization loss that needs to be adjusted, depending on the dataset. Mostly, we have used the hyper-parameter settings suggested by the author or recommended in the literature, such as higher $\beta$ value in the $\beta$-TCVAE for better disentanglement (typically set to $\beta \ge 5$) \cite{TC_VAE_Neurips_2019, Dis_benchmark_ICML_2019}. However, for multiple methods, such as the $\beta$-TCVAE and AAE, we have empirically determined the strength of the regularization loss for different datasets as we could not find them in the literature. For instance, the hyper-parameter $\beta$ in the $\beta$-TCVAE is set to $\beta=2$ (the minimum value) for the MNIST, CelebA, and CIFAR10 datasets after trying a range of values $\beta \in \{2, 4, 6, 10\}$, as higher $\beta$ value is leading to poor reconstruction. We report the hyper-parameter settings of the competing methods in Table \ref{tab:hyper-parameters} for all the benchmark datasets. 

Initialization of the model parameters has ramifications on the performance of the methods. Thus, we run all methods with $5$ different seeds (producing different initialization) for the MNIST, CelebA, and CIFAR10 datasets. Different initialization helps demonstrate the robustness of a method and allows statistical evaluation of the model's performance. 

\begin{table}[t]
\centering
\caption{Encoder and decoder architectures used in all the methods for the MNIST, CelebA, and CIFAR10 datasets. The architecture used in this study for multiple datasets is obtained from the literature \cite{RAE, tolstikhin2017wasserstein}.} \label{tab:Encoder_decoder_NN_1}
\resizebox{\textwidth}{!}
{%
\begin{tabular}{c c c c c c c}
\hline
&&{MNIST}&&{CelebA}&&{CIFAR10}\\ \hline
{Encoder: }
&&$\x{} \in \R{}^{32 \times 32 \times 1}$ 
&&$\x{} \in \R{}^{64 \times 64 \times 3}$
&& $\x{} \in \R{}^{32 \times 32 \times 3}$ \\
&&$\textsc{Conv64} \rightarrow \textsc{BN} \rightarrow \textsc{ReLU}$ 
&&$\textsc{Conv64} \rightarrow \textsc{BN} \rightarrow \textsc{ReLU}$
&& $\textsc{Conv128} \rightarrow \textsc{BN} \rightarrow \textsc{ReLU}$ \\
&&$\textsc{Conv128} \rightarrow \textsc{BN} \rightarrow \textsc{ReLU}$ 
&&$\textsc{Conv128} \rightarrow \textsc{BN} \rightarrow \textsc{ReLU}$
&& $\textsc{Conv256} \rightarrow \textsc{BN} \rightarrow \textsc{ReLU}$ \\
&&$\textsc{Conv256} \rightarrow \textsc{BN} \rightarrow \textsc{ReLU}$ 
&&$\textsc{Conv256} \rightarrow \textsc{BN} \rightarrow \textsc{ReLU}$
&& $\textsc{Conv512} \rightarrow \textsc{BN} \rightarrow \textsc{ReLU}$ \\
&&$\textsc{Conv512} \rightarrow \textsc{BN} \rightarrow \textsc{ReLU}$ 
&&$\textsc{Conv512} \rightarrow \textsc{BN} \rightarrow \textsc{ReLU}$
&& $\textsc{Conv1024} \rightarrow \textsc{BN} \rightarrow \textsc{ReLU}$ \\
&&$\textsc{Flatten}_{2 \times 2 \times 512} \rightarrow \textsc{FC}_{k \times 16} \rightarrow \textsc{None}$ &&$\textsc{Flatten}_{4 \times 4 \times 512} \rightarrow \textsc{FC}_{k \times 64} \rightarrow \textsc{None}$
&& $\textsc{Flatten}_{2 \times 2 \times 1024} \rightarrow \textsc{FC}_{k \times 128} \rightarrow \textsc{None}$ \\ \hline
{Decoder: }&&$\z{} \in \R{}^{16} \rightarrow \textsc{FC}_{2 \times 2 \times 512}$ &&$\z{} \in \R{}^{64}  \rightarrow \textsc{FC}_{8 \times 8 \times 512}$&& $\z{} \in \R{}^{128}  \rightarrow \textsc{FC}_{8 \times 8 \times 1024}$ \\
&&$\textsc{TransConv256} \rightarrow \textsc{BN} \rightarrow \textsc{ReLU}$ &&$\textsc{TransConv256} \rightarrow \textsc{BN} \rightarrow \textsc{ReLU}$&& $\textsc{TransConv512} \rightarrow \textsc{BN} \rightarrow \textsc{ReLU}$ \\
&&$\textsc{TransConv128} \rightarrow \textsc{BN} \rightarrow \textsc{ReLU}$ &&$\textsc{TransConv128} \rightarrow \textsc{BN} \rightarrow \textsc{ReLU}$&& $\textsc{TransConv256} \rightarrow \textsc{BN} \rightarrow \textsc{ReLU}$ \\
&&$\textsc{TransConv64} \rightarrow \textsc{BN} \rightarrow \textsc{ReLU}$ &&$\textsc{TransConv64} \rightarrow \textsc{BN} \rightarrow \textsc{ReLU}$&& $\textsc{TransConv3} \rightarrow \textsc{Sigmoid}$ \\
&&$\textsc{TransConv1} \rightarrow \textsc{Sigmoid}$ &&$\textsc{TransConv3} \rightarrow \textsc{Tanh}$&& \\
\hline
\end{tabular}
}
\end{table}

\subsection{Network Architectures}\label{supp:exp_settings}
In the neural network architectures reported in Table \ref{tab:Encoder_decoder_NN_1}, \textsc{Conv}${n}$ and \textsc{TransConv}${n}$ define convolution and transpose convolution operation, respectively, with $n$ filters in the output. We have used $4 \times 4$ filters for all the datasets. The transpose convolution filters use a stride size of $2$ except for the last layer of the decoders used in the CelebA and CIFAR10 datasets. We represent the fully connected layers as $\textsc{FC}_{k \times n}$ with $k \times n$ nodes, where $k=1$ for all the methods, except the VAE and $\beta$-TCVAE that use $k=2$. Activation functions used in the networks are ReLU (\textsc{ReLU}), Leaky ReLU (\textsc{LReLU}), sigmoid (\textsc{Sigmoid}), and hyperbolic tangent (\textsc{Tanh}). Input is in the range $[0, 1]$ for all the datasets except CelebA, for which the input is mapped to the range $[-1, 1]$. We use the Adam optimizer in all experiments (learning rate set to $5e-04$) with a learning rate scheduler (ReduceLROnPlateau) that reduces the learning rate by $0.5$ if the validation loss does not improve for $5$ epochs. All the methods are trained for $50$, $50$, and $100$ epochs for the MNIST, CelebA, and CIFAR10 datasets, respectively, except the VAE and $\beta$-TCVAE, which we trained for 100 epochs for the MNIST dataset. We use a batch size of $100$ to train all the methods. Additional optimization details are reported in Table \ref{tab:hyper-parameters}.

\begin{table}[t]
\centering
\caption{Discriminator architectures used by the AAE for the MNIST, CelebA, and CIFAR10 datasets.} \label{tab:Discr_AAE}
\resizebox{\textwidth}{!}
{%
\begin{tabular}{c c c c c}
\hline
{MNIST}&&{CelebA}&&{CIFAR10}\\ \hline
$\z{} \in \R{}^{16} \rightarrow \textsc{FC}_{100} \rightarrow \textsc{LReLU}$ 
&& $\z{} \in \R{}^{64} \rightarrow \textsc{FC}_{1024} \rightarrow \textsc{LReLU}$ 
&& $\z{} \in \R{}^{128} \rightarrow \textsc{FC}_{1024} \rightarrow \textsc{LReLU}$ \\
$\textsc{FC}_{100} \rightarrow \textsc{LReLU}$ 
&& $\textsc{FC}_{4096} \rightarrow \textsc{LReLU}$ 
&& $\textsc{FC}_{4096} \rightarrow \textsc{LReLU}$ \\
$\textsc{FC}_{100} \rightarrow \textsc{LReLU}$ 
&& $\textsc{FC}_{1024} \rightarrow \textsc{LReLU}$ 
&& $\textsc{FC}_{1024} \rightarrow \textsc{LReLU}$ \\
$\textsc{FC}_{1} \rightarrow \textsc{None}$ 
&& $\textsc{FC}_{1} \rightarrow \textsc{None}$ 
&& $\textsc{FC}_{1} \rightarrow \textsc{None}$ \\
\hline
\end{tabular}
}
\end{table}

The discriminator architectures in the AAE use only fully connected layers $\textsc{FC}_{n}$ with $n$ nodes in a layer and are defined in Table \ref{tab:Discr_AAE}. The discriminator takes as input the latent encodings, $\z{}$, produced by the encoder and predicts the probability of a sample to be \emph{real/fake}. We use Leaky ReLU ($\textsc{LReLU}$) activation in all the layers except the final layer.

\begin{table*}[t]
    \centering
    \caption{MSE per pixel of the competing methods (each method is trained $5$ times on a dataset, initialized differently in every run) on the benchmark datasets (lower is better). The \textbf{best} score is in \textbf{bold}, and the \underline{second best} score is \underline{{underlined}}.} \label{tab:mse}
    {%
    \begin{tabular}{|c||c||c||c|}\hline
    Method & MNIST $(l=16) \downarrow$ & CelebA $(l=64) \downarrow$ & CIFAR10  $(l=128) \downarrow$ \\ [0.5ex] 
    \hline\hline
    VAE             & $0.0115 \pm 0.0001$ & $0.0214 \pm 0.0000$ & $0.0161 \pm 0.0001$ \\
    $\beta$-TCVAE   & $0.0181 \pm 0.0003$ & $0.0239 \pm 0.0001$ & $0.0205 \pm 0.0000 $ \\
    RAE             & $\mathbf{0.0030 \pm 0.0001}$ & $0.0201 \pm 0.0001$ & $\underline{0.0063 \pm 0.0000}$ \\
    AAE             & $0.0069 \pm 0.0008$ & $0.0234 \pm 0.0028$ & $0.0101 \pm 0.0012$ \\
    WAE             &$\underline{0.0041 \pm 0.0001}$ & $\underline{0.0199 \pm 0.0001}$ & $0.0074 \pm 0.0001$ \\
    AVAE            & $\underline{0.0041 \pm 0.0004}$ & $\mathbf{0.0198 \pm 0.0001}$ & $\mathbf{0.0062 \pm 0.0002}$ \\ [1.5ex] \hline
    \end{tabular}}
\end{table*}

\begin{figure*}[htbp]
    \centering
    \includegraphics[width=1.0\textwidth]{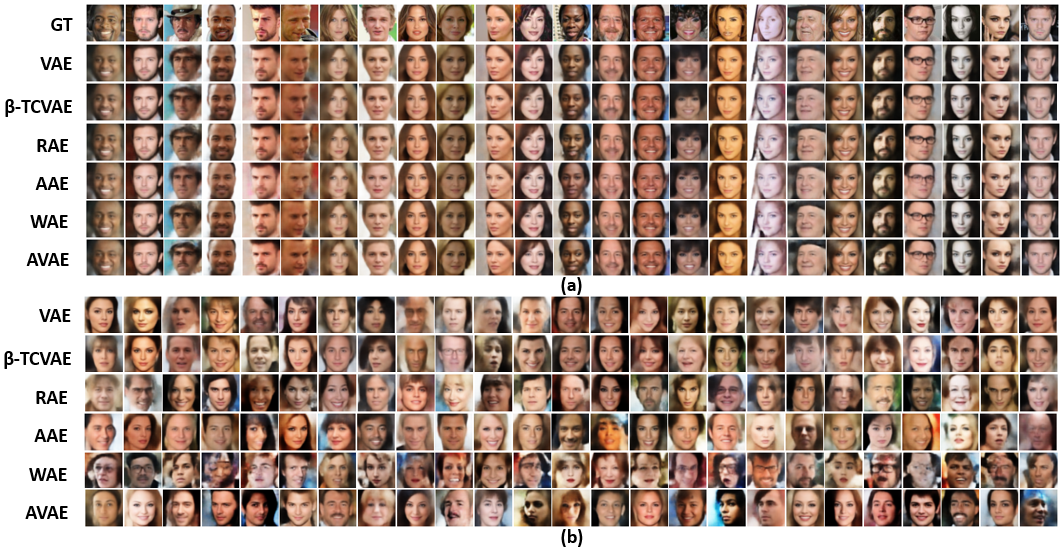}
    \caption{(a)Reconstructed and (b)generated examples of the CelebA dataset produced by the competing methods. \textbf{GT} represents the ground truth of the reconstructed examples.}
    \label{fig:CelebA_Images}
\end{figure*}

\subsection{Additional Results}\label{supp:add_results}

The reconstruction loss, i.e., the mean squared error (MSE) per pixel, is reported in Table \ref{tab:mse} for all the competing methods studied in this work. We compute the MSE on the held-out validation data of each benchmark dataset. We use the standard \emph{train, val}, and {\em test} split of the datasets. We use the same latent dimension ($l=16, 64,$ and $128$ for the MNIST, CelebA, and CIFAR10 datasets, respectively), encoder-decoder architecture (refer to Table \ref{tab:Encoder_decoder_NN_1}), and optimization strategies (refer to section \ref{supp:impl_details}) for all the competing methods to ensure a fair comparison. The hyper-parameter settings of each method for different datasets are reported in Table \ref{tab:hyper-parameters}. Reconstructed and generated images produced by different methods trained on the CelebA dataset are shown in Figure \ref{fig:CelebA_Images}.

\begin{figure}[t]
    \centering
    \includegraphics[width=1.0\textwidth]{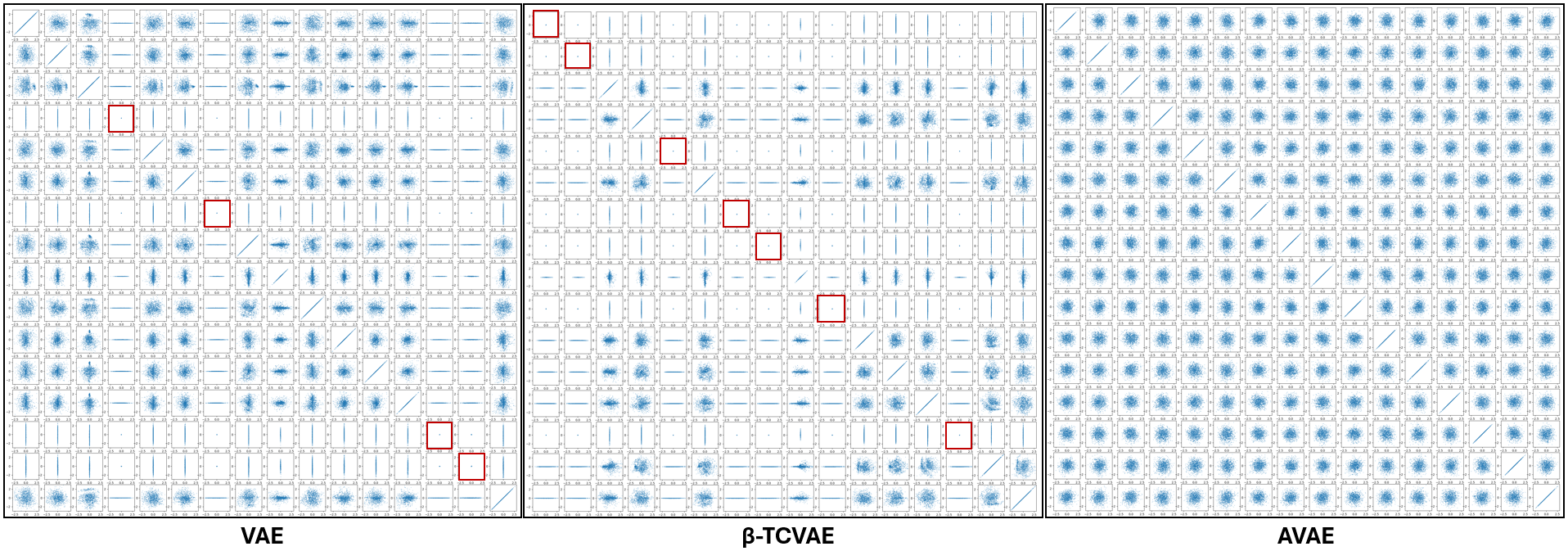}
    \caption{Pairwise scatter plot of the latent representations, $\z{} \in \R{}^{16}$, learned by different methods on the MNIST dataset. The cells highlighted in \textbf{red} color along the diagonal (for indices $(i, i)$) indicate the collapsing of the latent representations for the VAE and the $\beta$-TCVAE. Thus, we observe posterior collapse for multiple latent axes in the VAE ($4$ \textbf{latent axes}) and the $\beta$-TCVAE ($7$ \textbf{latent axes}). In addition, we observed clustering of the latent representations in the VAE and $\beta$-TCVAE in several latent axes pair, $(i, j) \text{ and } j \neq i$. These properties of the latent space manifest the failure of the VAE and $\beta$-TCVAE to match the aggregate posterior to the prior. In contrast, we do not encounter such properties in the latent representations produced by the AVAE, which help the AVAE achieve the best performance under different evaluation metrics on the MNIST dataset.}
    \label{fig:Posterior_collapse_MNIST}
\end{figure}

\begin{figure}[t]
    \centering
    \includegraphics[width=1.0\textwidth]{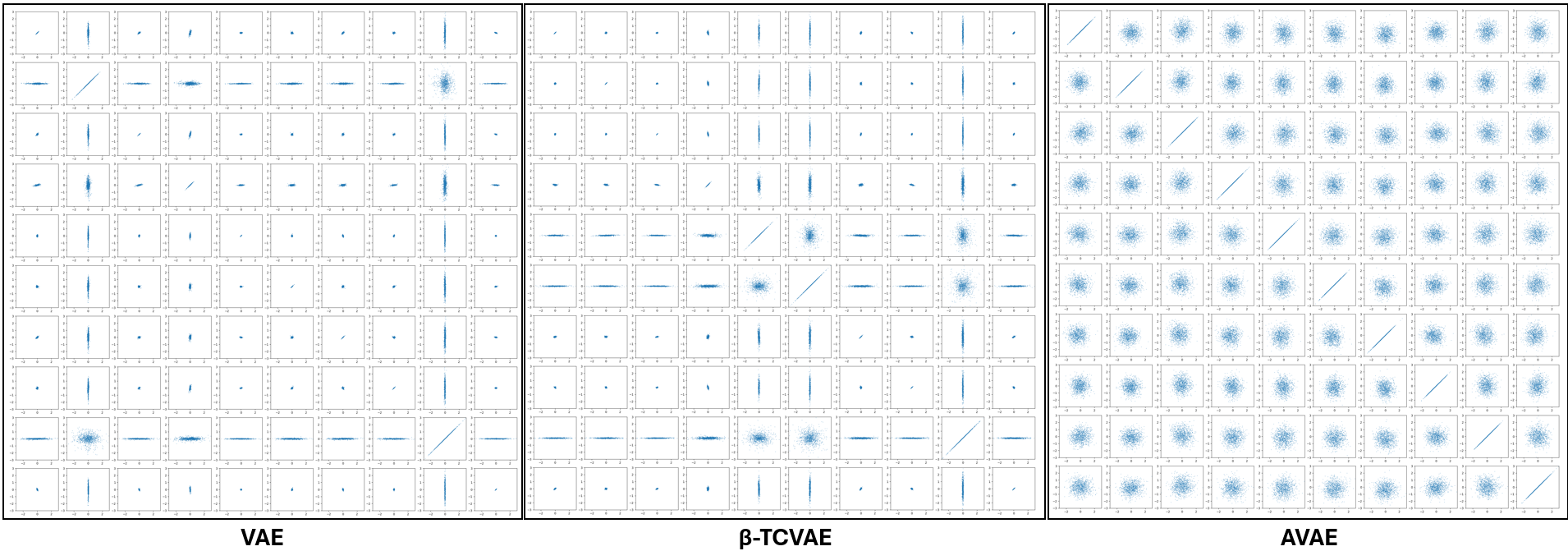}
    \caption{Pairwise scatter plot of the latent representations, $\z{} \in \R{}^{128}$, learned by different methods on the CIFAR10 dataset. For clarity, we use only the first ten ($10$) latent axes in the pairwise scatter plot. Similar to the MNIST dataset, we observe posterior collapse on the CIFAR10 dataset for the VAE and $\beta$-TCVAE. Most latent axes exhibit small variations in the mean representations produced by the VAE and $\beta$-TCVAE. In contrast, the AVAE does not suffer from such failures, and it matches the aggregate posterior distribution to the prior, even in a high-dimensional latent space, $l=128$.}
    \label{fig:Posterior_collapse_CIFAR10}
\end{figure}

The MSE of the competing methods in Table \ref{tab:mse} are comparable, except the MSE of the VAE and $\beta$-TCVAE on the MNIST and CIFAR10 datasets. We hypothesize that the high MSE on the MNIST and CIFAR10 datasets is possibly due to the posterior collapse in the VAE and $\beta$-TCVAE. To corroborate our hypothesis with evidence, we studied the structure in the latent space of the VAE and $\beta$-TCVAE using the pairwise scatter plot of the mean latent representations (not samples from the posterior distribution, $q_{\phi}(\z{} \mid \x{})$) learned on the MNIST and CIFAR10 datasets, as shown in Figure \ref{fig:Posterior_collapse_MNIST} and \ref{fig:Posterior_collapse_CIFAR10}, respectively. In this analysis, we use the deterministic encoding $\z{}_{i} = \mathbf{E_{\phi}}(\x{}_{i})$ as the latent representations for the AVAE. Typically for the MNIST dataset ($l=16$), the $i$-th row in a pairwise scatter plot represents the projection of the latent representations ($\in \R{}^{16}$) on the $i$-th latent axis, where each cell $(i, j)$ (with $j = \{1, 2, \dots, 15, 16\}$) shows the scatter plot for the latent axes pair $(i, j)$. Thus, the pairwise scatter plot size for the MNIST dataset is ${16} \times {16}$. The size of the pairwise scatter plot would change with the size of the latent space (indicated by $l$ in this work), such as ${128} \times {128}$ for the CIFAR10 using $l=128$. Visualizing the latent representations using the pairwise scatter plots presents another intuitive way of interpreting the properties of the latent space.

Suppose the mean representations learned by the VAE for the $i$-th latent axis have negligible or almost zero variance. In that case, it indicates the mapping of the observed samples to almost the same point on the latent axis by the encoder. This phenomenon is known as the posterior collapse \cite{Cycling_KL_annealing_VAE_NAACL_2019, posterior_collapse_delta_VAE_ICLR_2019} and we can observe this phenomenon along the diagonal (for indices $(i, i)$) in the pairwise scatter plot. For the VAE trained on the MNIST dataset, we encounter the posterior collapse in $4$ latent axes indicated by cells highlighted in \textbf{red} color along the diagonal in Figure \ref{fig:Posterior_collapse_MNIST}. The posterior distribution of the $\beta$-TCVAE collapsed on $7$ \textbf{latent axes} for the MNIST dataset. The collapsed dimensions in the VAE and $\beta$-TCVAE do not encode any features of the observed data, and thus, they do not pass any information to the decoder that can be used in the reconstruction of the observed data. The number of collapsed dimensions reduces the bottleneck capacity of a DLVM, resulting in higher reconstruction loss as observed in Table \ref{tab:mse} for the VAE and the $\beta$-TCVAE. Moreover, collapsed dimensions in a DLVM adversely affect its ability to learn the data distribution, leading to poor FID, precision, and recall scores (reported in Table 2 of the main paper). The number of collapsed dimensions in the VAE ($4$) and $\beta$-TCVAE ($7$) trained on the MNIST dataset are consistent across $5$ runs (model parameters are initialized differently in every run).

Furthermore, we observe the presence of holes or clusters in the latent space of the VAE and $\beta$-TCVAE trained on the MNIST dataset for several latent axes pair, $(i, j) \text{ and } j \neq i$. Besides the mMDS plot (reported in Figure 1 of the main paper), the pairwise scatter plot offers another way of visualizing the holes in the latent space of the VAE that results in low entropy scores for the VAE and $\beta$-TCVAE due to distribution mismatch. In contrast, we do not observe posterior collapse and holes in the latent space of the AVAE (manifested by uncorrelated latent representations for all latent axes pair, $(i, j) \text{ and } j \neq i$). This demonstrates the close matching of the aggregate posterior, $q_{\phi}$, to the prior in the AVAE and justifies the performance of the AVAE under different evaluation metrics.

We repeat the same analysis for the CFAR10 dataset. However, for brevity, we report the pairwise scatter plot of \emph{only the first \textbf{ten}} ($10$) latent axes of different models trained on the CIFAR10 dataset in Figure \ref{fig:Posterior_collapse_CIFAR10}. Similar to the MNIST dataset, we observe posterior collapse in the majority of the latent axes (out of $10$), both in the VAE and $\beta$-TCVAE. However, the AVAE does not suffer from this phenomenon. Moreover, there is no indication of holes in the latent space of the AVAE. Such properties of the distribution in the latent space help the AVAE achieve the best performance under different evaluation metrics on the CIFAR10 dataset.